\documentclass[letterpaper]{article} 
\usepackage[submission]{aaai2026}  
\usepackage{times}  
\usepackage{helvet}  
\usepackage{courier}  
\usepackage[hyphens]{url}  
\usepackage{amsmath}
\usepackage{graphicx} 
\usepackage{natbib}  
\usepackage{caption} 
\usepackage{algorithm}
\usepackage{algorithmic}
\usepackage{tabularx}
\usepackage{booktabs}
\usepackage{multirow}
\usepackage{amssymb}
\usepackage{newfloat}
\usepackage{listings}
\DeclareCaptionStyle{ruled}{labelfont=normalfont,labelsep=colon,strut=off} 
\floatstyle{ruled}
\newfloat{listing}{tb}{lst}{}
\floatname{listing}{Listing}
\title{Bi-level Meta-Policy Control for Dynamic Uncertainty Calibration in Evidential Deep Learning}
\author {
    Zhen Yang,
    Yansong Ma,
    Lei Chen
}

\usepackage{bibentry}

\begin{document}

\maketitle

\begin{abstract}
Traditional Evidence Deep Learning (EDL) methods rely on static hyperparameter for uncertainty calibration, limiting their adaptability in dynamic data distributions, which results in poor calibration and generalization in high-risk decision-making tasks. To address this limitation, we propose the Meta-Policy Controller (MPC), a dynamic meta-learning framework that adjusts the KL divergence coefficient and Dirichlet prior strengths for optimal uncertainty modeling. Specifically, MPC employs a bi-level optimization approach: in the inner loop, model parameters are updated through a dynamically configured loss function that adapts to the current training state; in the outer loop, a policy network optimizes the KL divergence coefficient and class-specific Dirichlet prior strengths based on multi-objective rewards balancing prediction accuracy and uncertainty quality. Unlike previous methods with fixed priors, our learnable Dirichlet prior enables flexible adaptation to class distributions and training dynamics. Extensive experimental results show that MPC significantly enhances the reliability and calibration of model predictions across various tasks, improving uncertainty calibration, prediction accuracy, and performance retention after confidence-based sample rejection. 
\end{abstract}


\section{Introduction}

In recent years, as deep learning technology has been widely applied in high-risk fields such as medical diagnosis and autonomous driving, the reliability and safety of model predictions have received unprecedented attention. In these real-world application scenarios\citep{kendall2017uncertainties, gawlikowski2023survey}, in addition to pursuing higher prediction accuracy, it is also considered highly important and necessary to accurately evaluate the model’s confidence in its own predictions—namely, uncertainty quantification (UQ). Effective uncertainty quantification not only improves the interpretability of models, helping developers and users understand the basis for model decisions, but also provides strong support for risk avoidance and decision-making, thus laying a solid foundation for the overall safety of systems and the trustworthiness of decisions.\par
\noindent
Among various UQ methods, by introducing the Dirichlet distribution to model class probabilities, EDL\citep{sensoy2018evidential} allows the model to output both classification results and corresponding uncertainty estimates in a single forward pass~\citep{ovadia2019trust, lakshminarayanan2017simple}, greatly improving inference efficiency and usability in practice. However, existing EDL methods still face some obvious limitations in real-world deployment. Specifically, During training, EDL models typically require the prior specification of key hyperparameters\citep{RevisitingEDL2024, radev2021amortized} such as the KL coefficient and Dirichlet prior to balance predictive accuracy and uncertainty representation. However, the optimal values of these hyperparameters are highly sensitive to the characteristics of the dataset, the nature of the task, and the particular phase of training. In practical scenarios\citep{xu2022uncertaintyaware, ancha2024deep}, researchers often need to engage in extensive trial-and-error and draw upon considerable experience to manually tune these parameters for different conditions. This heavy reliance on manual adjustment not only raises the barrier for model deployment and application, but also makes it difficult for EDL approaches to achieve efficient adaptation and generalization in the face of distributional shifts, task transitions, or continual learning settings\citep{guo2017calibration}.\par
\noindent
\begin{figure*}[htbp]
\centering
\includegraphics[width=0.95\linewidth]{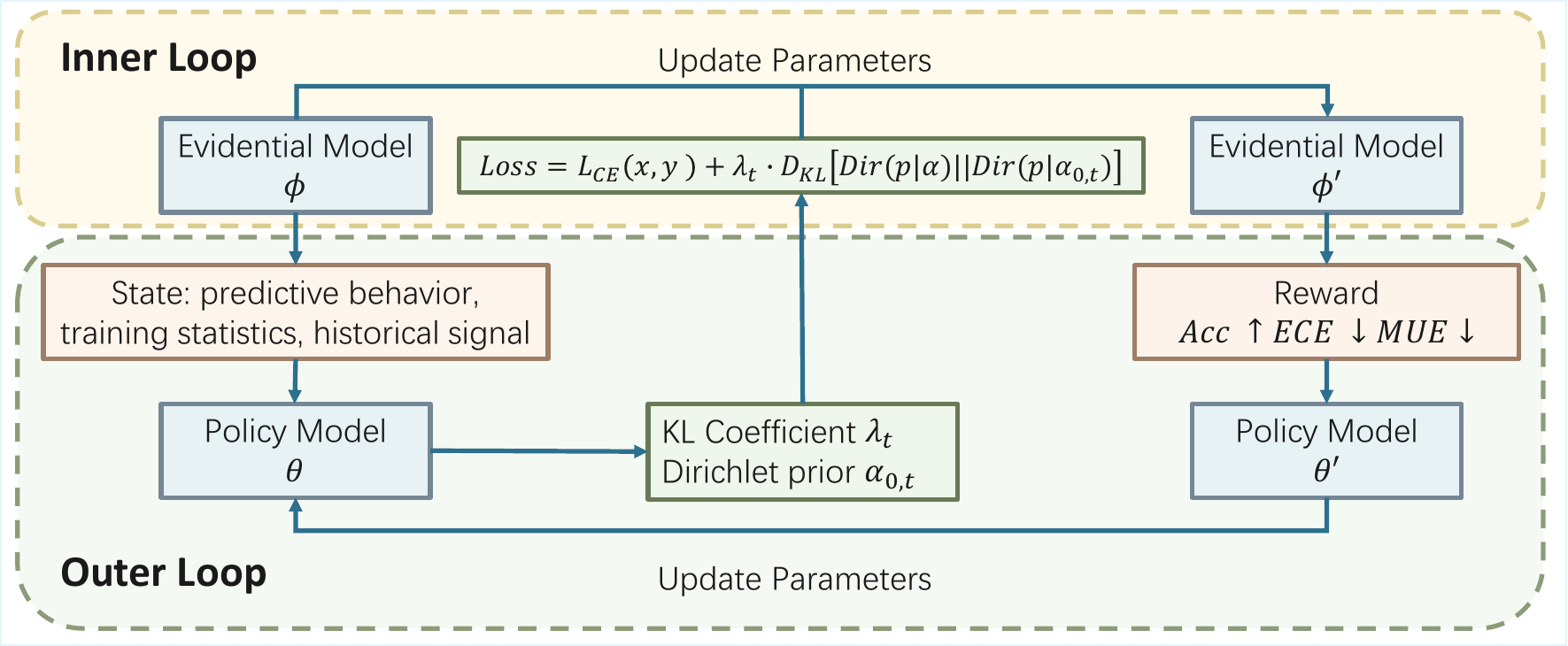}
\caption{Overview of our bilevel Meta-Policy Controller (MPC) framework. The inner loop updates the evidential model using a loss constructed with dynamically selected KL coefficient $\lambda_t$ and Dirichlet prior $\alpha_{0,t}$ from the policy. The outer loop updates the policy using reward feedback calculated from Accuracy, ECE, and MUE metrics.$\phi$,$\theta$ represents the model parameters respectively}
\label{fig:overview}
\end{figure*}
\noindent
Recent studies have attempted to introduce meta-learning frameworks that leverage feedback from the training process to dynamically adjust model behavior. Although meta-learning~\citep{kim2024semantic} has shown progress in multi-task learning and adaptive loss weighting, there remains a lack of dedicated dynamic regularization control mechanisms for uncertainty-aware deep learning models. Moreover, these approaches are difficult to apply directly to our core challenge, as some optimization targets---such as Expected Calibration Error (ECE) and Misclassification Uncertainty Error (MUE)---are non-differentiable, making real-time adjustment of key EDL hyperparameters challenging. To address this, we propose a meta-policy learning-based dynamic uncertainty calibration method tailored for EDL models. We design a state-aware policy network that dynamically adjusts the KL coefficient and Dirichlet prior based on the model’s current training status. Specifically, inspired by policy gradient methods in reinforcement learning, we employ a bi-level optimization framework with multiple reward signals to guide policy optimization. This enables flexible and intelligent hyperparameter adjustment, allowing the model to continually improve uncertainty estimation and prediction reliability across varying training phases and data distributions.\par

\vspace{0.5em}
\noindent
\textbf{The main contributions of this work are as follows:}
\begin{itemize}
    \item We present a novel framework that incorporates meta-learning into Evidential Deep Learning (EDL), enabling the dynamic regulation of uncertainty-aware loss functions through a learnable policy network.
    \item We design a state-aware policy network that adaptively adjusts both the KL regularization coefficient \(\lambda_t\) and the Dirichlet prior parameter \(\alpha_{0,t}\) in accordance with predictive performance and evolving training dynamics.
    \item We introduce, for the first time, a learnable and adaptive Dirichlet prior in EDL, replacing the conventional fixed uniform prior. This allows the model to flexibly update its prior belief strength across classes in response to training processes and distributional shifts, thereby enhancing uncertainty quantification.
\end{itemize}

\section{Backgrounds and Related Works}

\subsection{Problem Formulation and Limitations of Fixed Evidential Deep Learning}

Evidential Deep Learning (EDL)~\cite{Sensoy2018EDL} expresses predictive uncertainty by modeling outputs as a Dirichlet distribution parameterized by an evidence vector $e$. The total loss typically consists of two components: (i) a supervised data-fitting term, such as the mean squared error (MSE) between predicted and target soft labels; and (ii) a regularization term that penalizes deviation from a fixed Dirichlet prior $\alpha_0$ (usually chosen as a uniform distribution). The latter is commonly represented by the Kullback-Leibler (KL) divergence, scaled by a coefficient $\lambda$. The overall training objective is given by:
\begin{equation}
    \mathcal{L}_{\text{EDL}} = \mathcal{L}_{\text{fit}} + \lambda \cdot D_{\text{KL}}[\text{Dir}(\alpha) \,\|\, \text{Dir}(\alpha_0)].
\end{equation}
However, existing approaches often adopt static values\citep{shang2023relative} for $\lambda$ and $\alpha_0$, which are typically determined through cross-validation. This rigid design leads to several limitations:\par
\noindent
-\textbf{Training-phase mismatch:} The optimal value of $\lambda$ is inherently dynamic, evolving with the changing uncertainty and loss structure throughout the training process~\citep{nan2023region}.\par
\noindent
-\textbf{Data-distribution mismatch:} A fixed prior $\alpha_0$ fails to adapt to shifts in class distribution, label noise, or domain discrepancies.\par
\noindent
-\textbf{Coupling sensitivity:} Improper selection of $\lambda$ or $\alpha_0$ can simultaneously degrade both predictive performance and uncertainty calibration\citep{thiagarajan2022training}.\par
\noindent
\subsection{Evidential Deep Learning and Its Extensions}

Regularized Evidential Learning (RED)~\citep{zhang2023evcenter} improves the performance of EDL~\citep{sensoy2018evidential} by disentangling the necessary and unnecessary components of KL regularization, thereby enhancing the model's calibration and robustness. However, the fixed design of KL regularization may lead to over-smoothed belief distributions or fail to penalize overconfident incorrect predictions effectively~\citep{akbarifar2025stroke}. Recent surveys~\citep{gawlikowski2023survey, radev2021amortized} have summarized the advantages of EDL while highlighting its limitations, such as sensitivity to hyperparameters (e.g., the KL coefficient \(\lambda\) and the Dirichlet prior \(\alpha_0\)) and the trade-off between predictive accuracy and uncertainty quantification. These limitations are particularly pronounced in high-stakes scenarios such as medical diagnosis and autonomous driving~\citep{siebert2023uncertainty}.\par
\noindent
To address some of these challenges, several extensions of EDL have been proposed. For instance, Zhao et al.~\citep{zhao2019quantifying} introduced a regularization scheme that better constrains uncertainty in classification. Hu et al.~\citep{hu2021multidimensional} and Pandey \& Yu~\citep{pandey2022multidimensional} extended EDL to capture multidimensional and task-aware uncertainty, enabling more fine-grained belief representations. In open-set or unknown class scenarios, Nagahama~\citep{nagahama2023learning} proposed a method for modeling class-agnostic uncertainty, improving generalization to unseen data. Further, Zhang et al.~\citep{zhang2024revisiting} explored the interplay between domain shift and uncertainty modeling in active domain transfer settings, revealing the critical importance of calibrated belief estimation for domain-robust EDL.\par
\noindent
\subsection{Meta-Learning for Dynamic Loss Adjustment}

Meta-learning, or "learning to learn," is an emerging paradigm that dynamically adjusts model behavior based on feedback during the learning process. In recent years, strategy-based meta-learning frameworks such as MLAH~\citep{kim2024semantic} and NestedMAML~\citep{killamsetty2022nested} have demonstrated significant progress, showing that outer-loop policies can optimize inner-loop parameters, such as loss weighting and architecture selection. These methods leverage meta-gradients or reinforcement learning techniques to dynamically adjust learning objectives, thereby improving generalization and robustness.\par
\noindent
Unlike these methods, some studies attempt to improve uncertainty quantification performance by dynamically adjusting hyperparameters~\citep{wei2025openset}, such as KL coefficients or prior strengths. For example, the GradNorm method~\citep{chen2018gradnorm} dynamically balances loss weights in multi-task learning to address inconsistent learning speeds across tasks. However, these methods typically focus on dynamic trade-offs between task losses rather than addressing the specific needs of uncertainty quantification. Additionally, recent research~\citep{mansoori2023trace} on adaptive loss scaling has proposed a regularization adjustment mechanism based on training signals, but it primarily targets single calibration objectives and fails to simultaneously account for both classification performance and uncertainty quantification. More recently,~\citet{ramaswamy2023adaclip} proposed AdaClip, a task-aware loss modulation method that dynamically clips and scales loss contributions based on uncertainty and gradient variance, further underscoring the potential of adaptive loss formulations in uncertain settings.

\section{Method}
\subsection{Framework Overview}

We reformulate the conventional Evidential Deep Learning (EDL) training problem into a bilevel optimization framework by introducing a policy network $\pi_\theta$ that dynamically adjusts the KL coefficient and Dirichlet prior according to the current training state. The policy network takes as input the predictive behavior and training statistics of the model, and outputs values used to construct a customized EDL loss function for optimizing the backbone model.\par
\noindent
As shown in the Figure\ref{fig:overview}, to achieve joint optimization of the backbone model and the meta-policy, we adopt a two-timescale architecture: the backbone parameters are updated at a fast timescale based on the adaptive loss function, while the policy parameters are updated at a slower timescale using cumulative rewards that integrate model performance and uncertainty evaluation. This mechanism enables dynamic adaptation to data distribution shifts during training, effectively enhancing the model's generalization and uncertainty modeling capabilities. We have provided detailed pseudocode of the algorithm in the supplementary materials.

\subsection{State Representation and Policy Network Design}

We design a state-aware policy network \(\pi_\theta\) to adaptively adjust the KL coefficient and Dirichlet prior strength based on the current model state. Since a single indicator cannot fully describe the complex status of the model, the state vector \(s_t\) integrates three key types of information: (1) the model's immediate prediction performance, such as the current batch accuracy and mean output evidence, which reflect the model's confidence and discrimination ability; (2) training process statistics, including current loss and epoch index, which indicate the training stage and convergence progress; and (3) historical dynamic signals, such as the moving averages of the KL coefficient and validation accuracy, which capture long-term trends and fluctuations during training. This design enables the policy network to consider both the model's current performance and its historical evolution when making decisions, thus achieving more precise and stable adaptive loss configuration.

\paragraph{Network Architecture.}
The policy network consists of a shared multilayer perceptron that maps the input state \( s_t \) to a hidden representation \( h_t \in \mathbb{R}^{64} \):
\begin{equation}
h_t = \text{ReLU}(W_2 \cdot \text{ReLU}(W_1 \cdot s_t + b_1) + b_2).
\end{equation}
Based on \( h_t \), two separate output heads are used:\par
\noindent
\textbf{KL Coefficient Head} to produce \( \lambda_t \):
\begin{equation}
    \lambda_t = 10 \cdot \text{Sigmoid}(W_\lambda^\top h_t + b_\lambda),
\end{equation}
where \( \lambda_t \in (0, 10) \) is a small constant to avoid numerical instability.\par
\noindent
\textbf{Adaptive Prior Head} to produce a class-wise Dirichlet prior \( \alpha_{0,t} \in \mathbb{R}^{K}_+ \):
\begin{equation}
    \alpha_{0,t} = \text{Sigmoid}(W_{\alpha}^\top h_t + b_k) + 1, 
\end{equation}
The additive constant ensures that \( \alpha_{0,t}^{(k)} \geq 1 \), preserving the convexity of the KL divergence term.

\paragraph{State-Aware Loss Configuration.}
The policy outputs are used to configure the evidence-based training loss for the backbone model:
\begin{equation}
\begin{aligned}
\mathcal{L}_{\text{EDL}} &= \mathcal{L}_{\text{CE}}(\hat{y}, y) + \lambda \cdot D_{\text{KL}}\left[\text{Dir}(\alpha) \parallel \text{Dir}(\alpha_0)\right], \\
\end{aligned}
\end{equation}
where \( \alpha_0 = f_\phi(x_t) + 1 \) is the evidence output of the backbone model.

\noindent
This modular design enables the policy network to learn a lightweight yet expressive mapping from model states to uncertainty-aware loss configurations, effectively coordinating the trade-off between prediction accuracy and calibrated uncertainty. The detailed theoretical justification can be found in Appendix A.

\subsection{Meta-Policy Optimization}

We design a multi-objective reward mechanism for the policy network $\pi_\theta$ to guide it in learning regularization behaviors that benefit model generalization and uncertainty modeling. During training, the reward at each step is defined as
\begin{equation}
    R_t =  \Delta\text{ACC}_t - \beta_1 \cdot \Delta\text{ECE}_t - \beta_2 \cdot \Delta\text{MUE}_t,
\end{equation}
where $\text{ACC}_t$ denotes the batch-level classification accuracy, $\text{ECE}_t$ is the Expected Calibration Error that quantifies the mismatch between confidence and correctness, and $\text{MUE}_t$ measures the alignment between predictive uncertainty and actual misclassification. The coefficients $\beta_1$ and $\beta_2$ control the penalty weights on calibration and uncertainty misalignment, respectively, thereby encouraging the policy to pursue high accuracy while maintaining reliable uncertainty estimation.\par
\noindent
In our experiments, we set $\beta_1 = \beta_2 = 1$ across all datasets and tasks, which we found to be effective and stable without additional tuning. This default configuration provides balanced gradients among the three reward components. For practitioners aiming to emphasize either calibration or uncertainty more strongly, these coefficients can be manually adjusted to reflect task-specific priorities (e.g., increasing $\beta_1$ in safety-critical applications that require well-calibrated predictions).

\paragraph{Metric Definitions.}
\label{metric_def}

We now provide the formal definitions of the metrics used in our reward function:\par
\noindent
\textbf{(1) Accuracy (ACC):} Accuracy measures the overall classification correctness:
\begin{equation}
    \text{ACC} = \frac{1}{n} \sum_{i=1}^{n} 1(\hat{y}_i = y_i),
\end{equation}
where $\hat{y}_i$ and $y_i$ denote the predicted and true labels for sample $i$, and $n$ is the number of samples.\par
\noindent
\textbf{(2) Expected Calibration Error (ECE):} ECE quantifies the discrepancy between confidence and accuracy via binning:
\begin{equation}
    \text{ECE} = \sum_{m=1}^{M} \frac{|B_m|}{n} \left| \text{acc}(B_m) - \text{conf}(B_m) \right|,
\end{equation}
where $B_m$ is the set of predictions falling into the $m$-th confidence bin, and $\text{acc}(B_m), \text{conf}(B_m)$ denote the accuracy and mean confidence in that bin.\par
\noindent
\textbf{(3) Misclassification Uncertainty Error (MUE):} MUE evaluates whether predictive uncertainty $U_i$ aligns with prediction correctness:
\begin{equation}
\begin{split}
\text{MUE}(\tau) &= \textstyle\frac{1}{2} \left( \textstyle\frac{1}{|D_c|} \sum_{i \in D_c} 1(U_i > \tau) \right. \\
&\quad \left. + \textstyle\frac{1}{|D_i|} \sum_{i \in D_i} 1(U_i \leq \tau) \right)
\end{split}
\end{equation}
where $D_c$ and $D_i$ are the sets of correctly and incorrectly predicted samples respectively, and $\tau$ is a threshold selected to minimize MUE.\par
\noindent
Our overall framework implements meta-learning via alternating optimization: in the inner loop, model parameters are updated as
\begin{equation}
    \phi \leftarrow \phi - \eta \nabla_\phi L_{\text{EDL}}(\phi; \pi_\theta(s_t)),
\end{equation}
In the outer loop, we optimize the policy network parameters $\theta$ using the policy gradient method. Specifically, following the REINFORCE algorithm, the policy network samples an action (i.e., the hyperparameter configuration) based on the current state $s_t$, and updates its parameters according to the received reward $R_t$. The parameter update rule is given by:
\begin{equation}
    \theta \leftarrow \theta + \eta' \nabla_\theta \log \pi_\theta(a_t|s_t) R_t
\end{equation}
where $\pi_\theta(a_t|s_t)$ denotes the probability of the policy network selecting action $a_t$ (i.e., the configuration of the KL coefficient and Dirichlet prior) given state $s_t$, $R_t$ is the observed reward, and $b_t$ is a baseline to reduce variance. This approach allows the policy network to efficiently learn how to adjust the regularization hyperparameters through sampling and reward feedback.We have improved the detailed training algorithm in Appendix B.

\subsection{Theoretical Analysis of the Meta-Policy Controller}

In this section, we analyze the regret of the Meta-Policy Controller (MPC) from the perspective of online optimization, proving its convergence under the deterministic policy setting. Specifically, at each training step $t$, the policy network $\pi_\theta(s_t)$ generates hyperparameter configurations $a_t = (\lambda_t, \alpha_{0,t})$, which are optimized based on the current training loss $L_t(a_t) = L_{\text{EDL}}(\cdot; \lambda_t, \alpha_{0,t})$. Our objective is to minimize the cumulative regret:
\[
R_T = \sum_{t=1}^T L_t(a_t) - \min_{a \in \mathcal{A}} \sum_{t=1}^T L_t(a),
\]
where $\mathcal{A}$ denotes the feasible space of hyperparameters.

\subsubsection{Theoretical Setup and Assumptions}
We adopt the standard online convex optimization framework with the following assumptions:

\begin{enumerate}
    \item \textbf{Convexity and Lipschitz Continuity}: The loss function $L_t(a)$ is convex with respect to the hyperparameters $a$ and satisfies $G$-Lipschitz continuity, i.e., $\|\nabla L_t(a)\|_2 \leq G$.
    \item \textbf{Compact Action Space}: The hyperparameter space $\mathcal{A} \subset \mathbb{R}^d$ is a compact convex set with bounded diameter $D = \max_{a, a' \in \mathcal{A}} \|a - a'\|_2$.
    \item \textbf{Gradient Update Mechanism}: The policy network $\pi_\theta$ is updated via Projected Gradient Descent (PGD) to minimize the cumulative loss.
\end{enumerate}

\subsubsection{Regret Bound Analysis}
Based on these assumptions, we prove that the MPC's cumulative regret grows sublinearly, ensuring asymptotic convergence to the optimal static hyperparameter configuration. Under Assumptions 1-3, if MPC updates the policy network with learning rate $\eta = \frac{D}{G \sqrt{T}}$, the cumulative regret satisfies:

\[
R_T \leq G D \sqrt{T}.
\]
\subsubsection{Proof}
Since $L_t(a)$ is convex and the policy network is updated via Online Gradient Descent (OGD), we can apply the classical result from Zinkevich (2003):

\[
R_T \leq \frac{D^2}{2\eta} + \frac{\eta G^2 T}{2}.
\]
Substituting the optimal learning rate $\eta = \frac{D}{G \sqrt{T}}$ yields:

\[
R_T \leq G D \sqrt{T}.
\]
This bound indicates that the average regret $\frac{R_T}{T}$ approaches zero as $T \to \infty$, demonstrating that MPC asymptotically approaches the performance of the optimal static hyperparameter configuration.\par
\noindent
The theoretical analysis demonstrates that MPC, while constraining the hyperparameter feasible region (e.g., $\lambda_t \in [0.01,10]$, $\alpha_{0,t} \in [1.0,100.0]$), can effectively control the gradient variation of the loss function (whose Lipschitz constant $G$ is naturally bounded by softplus/ReLU activations and prior constraints) through online gradient mechanisms for dynamic hyperparameter adjustment. By maintaining a sublinear regret bound ($R_T \leq GD\sqrt{T}$) that grows with training steps $T$, MPC is theoretically guaranteed to asymptotically converge to the optimal static configuration. This provides verifiable feasibility justification for dynamic hyperparameter optimization in deep reinforcement learning.

\begin{table}[t]
\centering
\caption{Comparison of EDL, RED and our MPC method across datasets.}
\label{tab:baseline_comparison}
\setlength{\tabcolsep}{4pt}
\begin{tabular}{l c c c c c}
\toprule
\textbf{Dataset} & \textbf{Method} & \textbf{ACC↑} & \textbf{ECE↓} & \textbf{MUE↓} & \textbf{RACC↑} \\
\midrule
\multirow{3}{*}{MNIST}
  & EDL & 99.10 & 0.196 & 0.054 & 100.00 \\
  & RED & 99.20 & 0.196 & 0.048 & 100.00 \\
  & MPC & 99.30 & 0.274 & 0.061 & 100.00 \\
\midrule
\multirow{3}{*}{CIFAR-10}
  & EDL & 59.70 & 0.131 & 0.294 & 81.30 \\
  & RED & 60.20 & 0.111 & 0.293 & 82.10 \\
  & MPC & 71.80 & 0.413 & 0.255 & 90.00 \\
\midrule
\multirow{3}{*}{SVHN}
  & EDL & 86.70 & 0.116 & 0.167 & 97.40 \\
  & RED & 86.80 & 0.118 & 0.164 & 97.40 \\
  & MPC & 91.30 & 0.409 & 0.153 & 98.40 \\
\bottomrule
\end{tabular}
\end{table}
\begin{table}[t]
\centering
\small
\caption{Impact of different reward designs on SVHN.}
\label{tab:reward_impact_svhn}
\begin{tabularx}{\linewidth}{l *{4}{>{\centering\arraybackslash}X}}  
\toprule
\textbf{Reward Design} & \textbf{ACC↑} & \textbf{ECE↓} & \textbf{MUE↓} & \textbf{RACC↑} \\
\midrule
ECE + MUE       & 90.50 & 0.374 & 0.166 & 98.10 \\
ACC + ECE       & 90.60 & 0.356 & 0.171 & 97.70 \\
ACC + MUE       & 90.90 & 0.447 & 0.149 & 98.40 \\
ACC + ECE + MUE & 90.80 & 0.438 & 0.154 & 98.20 \\
\bottomrule
\end{tabularx}
\end{table}

\begin{table}[htbp]
\centering
\small                 
\setlength{\tabcolsep}{3pt}
\caption{Ablation of state components on SVHN.}
\label{tab:state_ablation}
\begin{tabularx}{\linewidth}{l *{4}{>{\centering\arraybackslash}X}}
\toprule
\textbf{State Component Removed} &
\textbf{ACC↑} & \textbf{ECE↓} & \textbf{MUE↓} & \textbf{RACC↑} \\
\midrule
None (Full)     & 91.30 & 0.4091 & 0.1532 & 98.40 \\
acc             & 90.75 & 0.4321 & 0.1581 & 98.08 \\
mean\_evidence  & 90.99 & 0.4529 & 0.1506 & 98.46 \\
loss\_feat      & 90.86 & 0.4490 & 0.1516 & 98.31 \\
epoch\_feat     & 90.65 & 0.4347 & 0.1538 & 98.39 \\
val\_acc\_feat  & 90.65 & 0.4329 & 0.1530 & 98.27 \\
hist\_kl\_feat  & 90.68 & 0.4406 & 0.1549 & 98.26 \\
\bottomrule
\end{tabularx}
\end{table}
\begin{figure}[t]
\centering
\includegraphics[width=\linewidth ]{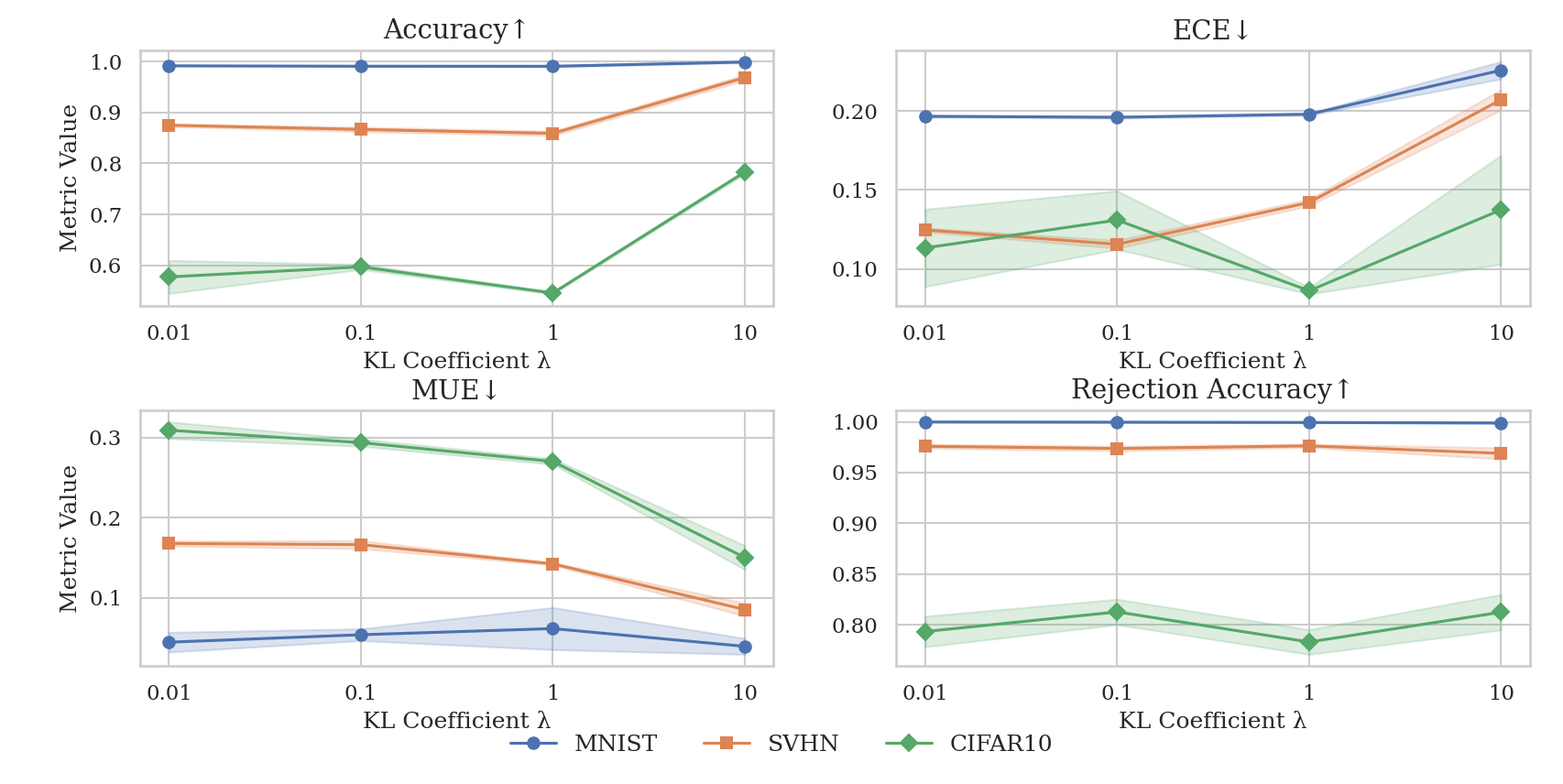}
\caption{Performance curves (mean and standard deviation) under different KL coefficients on MNIST, CIFAR-10, and SVHN datasets. Dynamic trends show dataset-specific preferences for KL settings, motivating adaptive strategies.}
\label{fig:kl_sensitivity_curve}

\end{figure}
\begin{table}[t]
\centering
\small
\caption{Performance and efficiency under different policy update intervals on SVHN.}
\label{tab:interval_time}
\begin{tabularx}{\linewidth}{l*{5}{>{\centering\arraybackslash}X}}
\toprule
\textbf{Method} & \textbf{ACC} & \textbf{ECE} & \textbf{MUE} & \textbf{RACC} & \textbf{Time(s)} \\
\midrule
EDL     & 86.70 & 0.116 & 0.167 & 97.40 & 12.58 \\
RED     & 86.80 & 0.118 & 0.164 & 97.40 & 11.84 \\
MPC-1   & 90.80 & 0.438 & 0.154 & 98.20 & 34.56 \\
MPC-2   & 91.00 & 0.426 & 0.152 & 98.30 & 23.54 \\
MPC-3   & 91.30 & 0.409 & 0.153 & 98.40 & 19.76 \\
MPC-5   & 91.10 & 0.389 & 0.161 & 98.30 & 14.71 \\
MPC-10  & 90.00 & 0.363 & 0.168 & 97.70 & 14.16 \\
\bottomrule
\end{tabularx}
\end{table}
\begin{table}[htbp]
\centering
\caption{Late-stage convergence statistics across datasets (20\% epochs)}
\label{tab:convergence_compact}
\begin{tabularx}{\linewidth}{l*{3}{>{\centering\arraybackslash}X}c}
\toprule
\textbf{Dataset} & \textbf{$\lambda$ CV } & \textbf{$\alpha_0$ CV } & \textbf{Acc CV } & \textbf{Stability} \\
\midrule
MNIST    & 13.50\% & 0.39\% & 0.10\% & \checkmark \\
SVHN     & 3.68\%  & 0.09\% & 0.44\% & \checkmark \\
CIFAR-10 & 1.22\%  & 0.03\% & 1.06\% & \checkmark \\
\bottomrule
\end{tabularx}
\end{table}

\section{Experiments}
\subsection{Experimental Setting}

To comprehensively evaluate the effectiveness of our proposed Meta-Policy Controller (MPC) framework, we conduct experiments across several benchmark datasets. For classification tasks on MNIST, CIFAR-10, and SVHN, all models adopt a ResNet18 backbone and an evidential output layer with ReLU evidence activation, trained using the Adam optimizer with a learning rate of \(1 \times 10^{-4}\). The training epochs are set to 50 for MNIST, 100 for SVHN, and 200 for CIFAR-10. The MPC itself utilizes shared layers with 128 and 64 output dimensions, ReLU activation, and is optimized using three-round delayed rewards where coefficients \(\beta_1\) and \(\beta_2\) in the reward function are both set to 1. We evaluate models on Accuracy (ACC), Expected Calibration Error (ECE), Misclassification Uncertainty Error (MUE), and Retained Accuracy (RACC), along with Retained Accuracy (RACC) under Confidence Calibration measures the classification accuracy on the subset of samples retained after rejecting those with uncertainty exceeding a specified threshold:
\begin{equation}
    \text{RACC} = \frac{1}{|D_{\text{retained}}|} \sum_{i \in D_{\text{retained}}} 1(\hat{y}_i = y_i),
\end{equation}
where $D_{\text{retained}} = \{i \mid U_i \leq \tau^* \}$ and $\tau^*$ is the MUE-optimal uncertainty threshold.  All results are averaged over three independent runs and conducted on an NVIDIA GeForce RTX 3070 Ti Laptop GPU.\par
\noindent
\noindent
\textbf{Experiment comparing with baseline} We begin by validating the core premise of our work. Traditional Evidential Deep Learning (EDL) methods rely on a static KL coefficient, which, as shown in Figure \ref{fig:kl_sensitivity_curve}, requires dataset-specific tuning and struggles to balance accuracy and calibration. This highlights the necessity of a dynamic adjustment mechanism. We therefore systematically compare MPC against mainstream EDL and its improved version, RED. Under an identical backbone, where the only difference is the regularization strategy, MPC demonstrates significant improvements across both classification accuracy and uncertainty calibration metrics (ECE, MUE), particularly on complex datasets like CIFAR-10 and SVHN. As detailed in Table~\ref{tab:baseline_comparison}, MPC not only enhances overall accuracy but also markedly reduces calibration and misclassification errors. Furthermore, its superior RACC score indicates that the uncertainty estimates produced by MPC are more discriminative, allowing for higher accuracy on samples retained after filtering by an optimal uncertainty threshold. This underscores MPC's ability to improve system reliability in real-world applications.More experiments about ECE can be found in the supplementary materials. \par
\noindent
\textbf{Ablation Study} Having established the superior performance of MPC, we conduct ablation studies to dissect the key components of its policy design. On the SVHN dataset, we first analyze the composition of the reward signal. As shown in Table~\ref{tab:reward_impact_svhn}, different reward combinations yield distinct trade-offs. Notably, the combination of \textit{ACC + MUE} achieves the best overall performance with the highest accuracy (90.90\%) and lowest MUE (0.149). The full combination of \textit{ACC + ECE + MUE} also delivers near-optimal results across all metrics, confirming that a holistic reward signal fosters robust and balanced optimization. Secondly, we examine the state representation by progressively removing features. The results in Table~\ref{tab:state_ablation} reveal that removing any single feature degrades performance, with the absence of the historical KL coefficient and validation accuracy having the most significant impact. This suggests that a rich, multi-faceted state encoding is essential for the policy network to effectively guide the training process.\par
\noindent

\begin{table*}[t]
\centering
\caption{OOD Uncertainty Estimation,Accuracy on Top-K confident samples}
\label{tab:uncertainty_evaluation}
 \resizebox{\textwidth}{!}{ 
\begin{tabular}{lccccccc}
\toprule
\textbf{Methood} & \textbf{Top 10\%} & \textbf{Top 20\%} & \textbf{Top 30\%} & \textbf{Top 50\%} & \textbf{Top 80\%} & \textbf{Top 100\%} & \textbf{OOD Reject Rate$\uparrow$(\%)} \\
\midrule
EDL & 95.90& 92.30& 88.50& 79.10& 66.70& 59.70& 78.00\\
RED & 96.60& 93.80& 89.40& 80.20& 67.80& 60.60& 78.00\\
MPC & 96.40& 95.50& 94.40& 90.40& 79.60& 70.90& 87.50\\
\bottomrule
\end{tabular}
}
\end{table*}

\begin{table*}[t]
\centering
\caption{Comparative results on CIFAR-10-LT (Accuracy values reported in percentage)}
\label{tab:main_results}
\resizebox{\textwidth}{!}{ 
\begin{tabular}{l|ccccccccccc|cccc}
\hline
\multirow{2}{*}{Method} & \multicolumn{10}{c}{Class-wise Accuracy$\uparrow$(\%)} & & \multirow{2}{*}{ECE$\downarrow$} & \multirow{2}{*}{MUE$\downarrow$} & \multirow{2}{*}{RACC $\uparrow$(\%)} \\
& 0 & 1 & 2 & 3 & 4 & 5 & 6 & 7 & 8 & 9 & Avg & & & \\
\hline
EDL & 95.1 & 93.5 & 73.7 & 56.2 & 53.4 & 43.0 & 40.3 & 46.0 & 9.1 & 0.9 & 51.1 & 8.8 & 36.2 & 63.4 \\
RED & 95.0 & 94.4 & 75.0 & 54.2 & 58.3 & 33.2 & 64.7 & 47.9 & 5.5 & 3.4 & 53.2 & 6.7 & 36.2 & 65.3 \\
MPC (Ours) & 93.8 & 90.6 & 56.2 & 56.6 & 61.2 & 26.2 & 78.0 & 44.7 & 47.3 & 36.5 & 64.1 & 49.4 & 28.5 & 84.4 \\
\hline
\end{tabular}
}
\end{table*}

\par\noindent
\textbf{Experiment on Time Cost and Training Strategy} Beyond performance, the practical viability of MPC hinges on its computational efficiency and training stability. We first analyze the trade-off between performance and overhead by varying the policy update interval on the SVHN dataset, benchmarked against EDL and RED. As detailed in Table~\ref{tab:interval_time}, the update frequency is a critical factor. Overly frequent updates (e.g., every epoch) introduce significant computational overhead and risk destabilizing the policy by forcing it to react to noisy, short-term fluctuations in the training state. Conversely, overly sparse updates (e.g., every 10 epochs) cause the policy to lag behind the main model's learning progress, resulting in stale, suboptimal hyperparameter adjustments that degrade overall accuracy and adaptability. An interval of 3 epochs emerges as the optimal balance, affording the policy a sufficiently stable state representation for robust decision-making while remaining agile enough to guide the training dynamics effectively. Crucially, even with a 5-epoch interval, MPC maintains competitive performance with only a marginal increase in training time over the non-adaptive baselines, confirming its practicality for real-world deployment.\par
\begin{figure}[t]
    \centering
    \includegraphics[width=\linewidth]{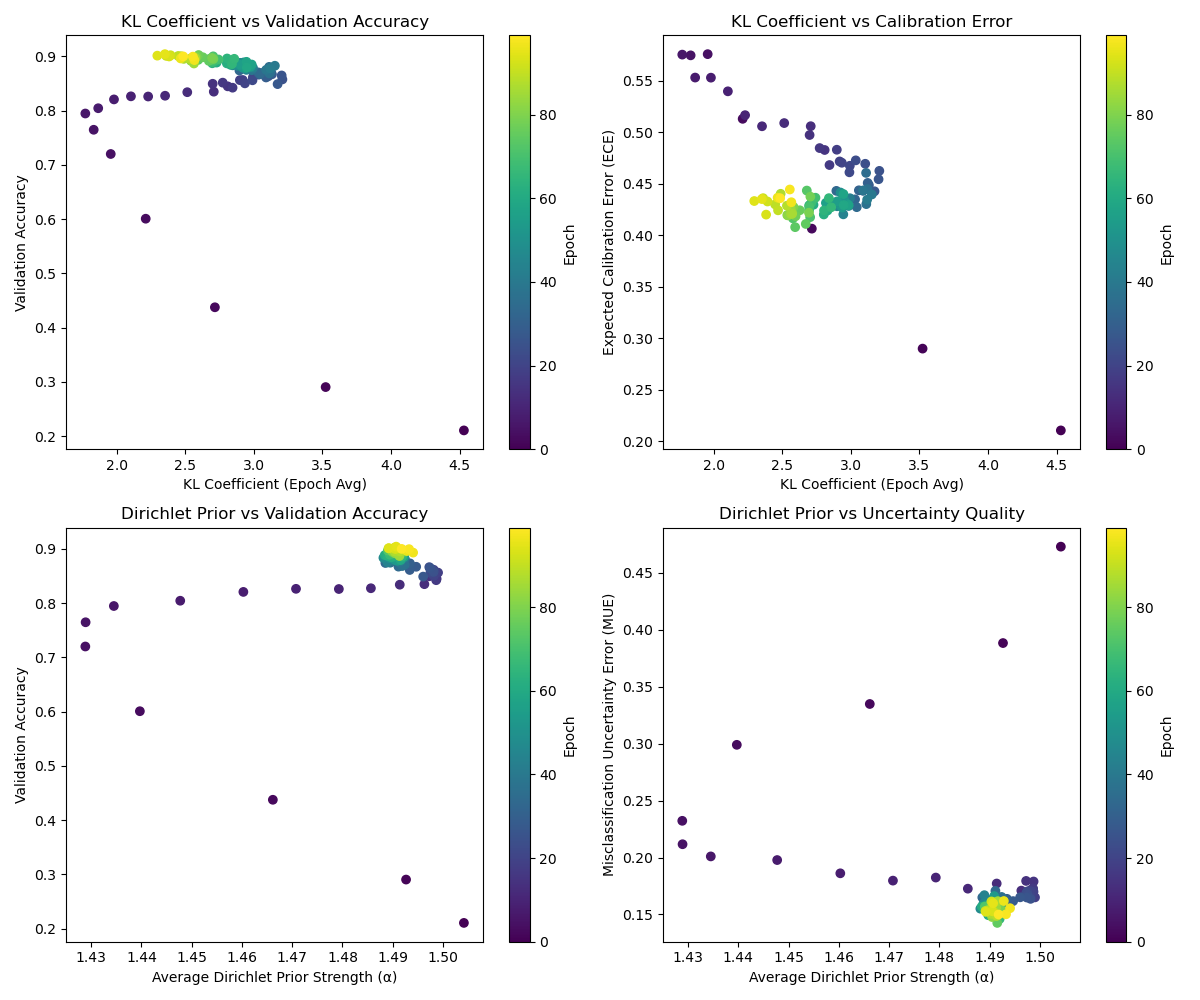}
    \caption{Correlation between hyperparameter dynamics (KL coefficient and Dirichlet prior strength) and performance metrics during training on SVHN. Each dot represents one epoch.}
    \label{fig:hyperparam_corr}
\end{figure}
\noindent
\textbf{Convergence analysis} Furthermore, a key concern for any dynamic method is whether its adaptive nature leads to erratic or unstable training. We address this by assessing the convergence and stability of MPC during the final 20\% of training epochs. By calculating the coefficient of variation (CV) of key hyperparameters and validation accuracy, we can quantify the training dynamics. The results, summarized in Table \ref{tab:convergence_compact} and visualized in Figure \ref{fig:hyperparam_corr}, provide strong evidence of stability. The KL coefficient $\lambda_t$ and Dirichlet prior $\alpha_0$,$t$ settle into a consistent regime, with remarkably low CVs (below 14\% and 0.4\%, respectively). Figure \ref{fig:hyperparam_corr} further illustrates the correlation between the dynamics of these hyperparameters (KL coefficient and Dirichlet prior strength) and performance metrics during training on SVHN, demonstrating a stable and predictable relationship. This indicates that the meta-policy is not oscillating but has converged to a stable control strategy tailored to the dataset. This stability is further reflected in the unwavering validation accuracy, which confirms that the converged policy successfully steers the model towards a robust and high-performing solution. Together, these findings demonstrate that MPC is not only efficient but also maintains reliable and predictable training dynamics, making it a trustworthy and practical framework.\par
\noindent
\par\noindent
\textbf{Experiments in OOD and Unbalanced Scenarios} To probe the limits of our framework's generalization, we subject it to more demanding scenarios that challenge standard models: out-of-distribution (OOD) detection and long-tailed recognition. First, in an OOD task designed to test the model's ability to handle unknown inputs, we use CIFAR-10 as in-distribution data and SVHN as unseen OOD data. The results in Table~\ref{tab:uncertainty_evaluation} show that MPC decisively outperforms EDL and RED. By dynamically tuning the KL regularization, MPC learns a more compact and precise representation of the in-distribution data manifold. This results in not only higher accuracy on high-confidence in-distribution samples but also a significantly higher rejection rate for OOD samples, whose feature representations naturally fall outside this learned manifold. This demonstrates a superior discriminative capability crucial for risk control in open-world environments. We then extend this robustness evaluation to severe data distribution shifts using the long-tailed CIFAR-10-LT dataset. As shown in Table~\ref{tab:main_results}, MPC again surpasses the baselines across all metrics. Its true strength is revealed in the substantial accuracy gains on the data-scarce tail classes. This confirms that MPC's adaptive policy effectively counteracts the tendency of models to become overconfident on head classes by dynamically adjusting the evidential prior and regularization, thereby forcing a more equitable allocation of learning capacity and yielding more reliable predictions across the entire class spectrum.\par
\noindent

\begin{table}[t]
\centering
\caption{Comparison of segmentation performance on the GlaS dataset (average over 3 runs).}
\label{tab:glas_segmentation}
\begin{tabularx}{\linewidth}{l*{3}{>{\centering\arraybackslash}X}} 
\toprule
\textbf{Method} & \textbf{IoU $\uparrow$} & \textbf{ECE $\downarrow$} & \textbf{MUE $\downarrow$} \\
\midrule
EDL & 0.733 & 0.127 & 0.311 \\
RED & 0.721 & 0.138 & 0.294 \\
MPC (Ours) & 0.784 & 0.255 & 0.234 \\
\bottomrule
\end{tabularx}
\end{table}

\textbf{Experiment in Medical Image Segmentation } Finally, to validate the cross-domain generalizability of our method in safety-critical applications, we apply it to medical image segmentation on the GlaS dataset. Using a U-Net backbone, we tasked the models with producing well-calibrated, pixel-wise predictions. The results in Table \ref{tab:glas_segmentation} show that MPC outperforms the baselines in both Intersection over Union (IoU) and MUE. Despite a slight increase in ECE, the significant improvement in MUE points to more reliable uncertainty estimation at the pixel level. This successful application to a complex segmentation task underscores the strong cross-domain adaptability and potential of MPC for high-stakes, real-world problems.

\section{Conclusion}

This study introduces a Meta-Policy Controller (MPC) based on reinforcement learning, designed to dynamically adjust the KL regularization coefficient and Dirichlet prior in evidential deep learning. By addressing the limitations of static regularization, MPC significantly enhances uncertainty calibration. Empirical results demonstrate that MPC not only improves predictive performance but also achieves a better alignment between classification accuracy and uncertainty estimation.

\noindent
Despite its effectiveness, MPC introduces additional training overhead due to the incorporation of a policy optimization module. Although delayed updates help alleviate this cost, the dynamic tuning mechanism increases implementation complexity compared to conventional static approaches.

\noindent
Notably, MPC exhibits strong robustness under distribution shifts, particularly in safety-critical applications such as medical diagnosis. Future work will explore how to better incorporate domain-specific knowledge into the policy learning process to establish more reliable and interpretable safety mechanisms.

\bibliography{refs}
\appendix
\section{Appendix}
\subsection{Algorithm}

To implement the meta-policy framework, we adopt a two-timescale alternating optimization strategy between the evidential learner and the policy controller. The inner loop trains the evidence-based model using dynamically selected KL coefficient $\lambda$ and Dirichlet prior $\alpha_0$. The outer loop updates the policy network using metric-based rewards derived from episodic evaluation.

\begin{algorithm}[H]
\caption{Meta-Policy Controlled Evidential Learning}
\label{alg:mpc}
\begin{algorithmic}[1]
    \REQUIRE Training data $\mathcal{D}$, meta-policy $\pi_\theta$, evidential model $f_\phi$
    \STATE Initialize policy parameters $\theta$ and model parameters $\phi$
    \WHILE{each outer iteration $e = 1, \dots, E$}
        \STATE Sample a training minibatch $\{(x_i, y_i)\}_{i=1}^B \sim \mathcal{D}$
        \STATE Compute evidence $e = f_\phi(x)$ and construct state $s = \text{GetState}(e, y, \cdots)$
        \STATE Generate actions $(\lambda, \alpha_0) \leftarrow \pi_\theta(s)$
        \STATE Compute EDL loss $\mathcal{L}_{\text{EDL}}(x, y; \lambda, \alpha_0)$
        \STATE Update model parameters: $\phi \leftarrow \phi - \eta_\phi \nabla_\phi \mathcal{L}_{\text{EDL}}$
        \IF{end of episode or scheduled evaluation}
            \STATE Evaluate validation metrics: $\text{ACC}_t$, $\text{ECE}_t$, $\text{MUE}_t$
            \STATE Compute reward: $R_t = \text{ACC}_t - \beta_1 \cdot \text{ECE}_t - \beta_2 \cdot \text{MUE}_t$
            \STATE Update policy: $\theta \leftarrow \theta + \eta_\theta \nabla_\theta \mathbb{E}_{\pi_\theta}[R_t]$
        \ENDIF
    \ENDWHILE
\end{algorithmic}
\end{algorithm}

\vspace{0.5em}
\noindent
\textbf{Remarks.}
To enhance efficiency and stability, we use shared feature encoders in the policy network. Empirical metric trends (e.g., accuracy mean evidence, history KL) are included in the policy state vector for informed adaptation.

\subsection*{Performance of EDL with Fixed KL Coefficients}

For completeness, we provide the detailed performance of standard EDL with fixed KL regularization strengths on three datasets: MNIST, SVHN, and CIFAR-10. While the main paper presents trends via line plots, Table~\ref{tab:edl_fixed_kl} reports the exact numerical values across four KL coefficients $\lambda \in \{0.01, 0.1, 1, 10\}$.

The results show that performance is highly sensitive to the choice of $\lambda$: overly small or large values can hurt either accuracy or calibration. For example, on CIFAR-10, larger $\lambda$ (e.g., 10) significantly degrades accuracy, while moderate values achieve better balance.

\begin{table}[htbp]
\centering
\caption{EDL performance under different fixed KL coefficients ($\lambda$).}
\label{tab:edl_fixed_kl}
\begin{tabular}{llcccc}
\toprule
\textbf{Dataset} & \textbf{Metric} & \textbf{0.01} & \textbf{0.1} & \textbf{1} & \textbf{10} \\
\midrule
\multirow{4}{*}{MNIST} 
& ACC $\uparrow$ & 0.9920 & 0.991 & 0.991 & 0.956 \\
& ECE $\downarrow$ & 0.197 & 0.196 & 0.198 & 0.226 \\
& MUE $\downarrow$ & 0.045 & 0.054 & 0.062 & 0.040 \\
& RACC $\uparrow$ & 1.000 & 1.000 & 0.999 & 0.999 \\
\midrule
\multirow{4}{*}{CIFAR10} 
& ACC $\uparrow$ & 0.577 & 0.597 & 0.545 & 0.473 \\
& ECE $\downarrow$ & 0.113 & 0.131 & 0.086 & 0.137 \\
& MUE $\downarrow$ & 0.309 & 0.294 & 0.270 & 0.151 \\
& RACC $\uparrow$ & 0.794 & 0.813 & 0.783 & 0.813 \\
\midrule
\multirow{4}{*}{SVHN} 
& ACC $\uparrow$ & 0.875 & 0.867 & 0.859 & 0.754 \\
& ECE $\downarrow$ & 0.113 & 0.116 & 0.142 & 0.207 \\
& MUE $\downarrow$ & 0.168 & 0.167 & 0.143 & 0.086 \\
& RACC $\uparrow$ & 0.976 & 0.974 & 0.976 & 0.969 \\
\bottomrule
\end{tabular}
\end{table}

\subsection{Additional Experiments}

\subsubsection*{Effect of $\beta_1$ on Calibration}

To investigate the influence of the reward coefficient $\beta_1$ in our Meta-Policy Controller (MPC), we conduct a sensitivity analysis on three datasets: MNIST, SVHN, and CIFAR-10. We vary $\beta_1$ in \{0.1, 1, 2.5, 5, 10\} while keeping $\beta_2 = 1$ fixed.

As shown in Table~\ref{tab:beta1_sweep}, increasing $\beta_1$ significantly improves the Expected Calibration Error (ECE) across all datasets. Notably, this calibration gain is achieved without compromising accuracy, misclassification uncertainty (MUE), or retained accuracy (RACC). In fact, moderate tuning (e.g., $\beta_1{=}5$ or $10$) leads to consistent improvements across all evaluation metrics, confirming that MPC's performance can be further enhanced through lightweight hyperparameter tuning.
\begin{table}[htbp]
\centering
\caption{Effect of $\beta_1$ on performance across datasets.}
\label{tab:beta1_sweep}
\resizebox{\columnwidth}{!}{
\begin{tabular}{llcccc}
\toprule
\textbf{Dataset} & \textbf{$\beta_1$} & \textbf{ACC $\uparrow$} & \textbf{ECE $\downarrow$} & \textbf{MUE $\downarrow$} & \textbf{RACC $\uparrow$} \\
\midrule
\multirow{7}{*}{MNIST} 
& 0.1 & 0.9927 & 0.290 & 0.054 & 0.9998 \\
& 1.0 & 0.9925 & 0.274 & 0.061 & 0.9996 \\
& 2.5 & 0.9922 & 0.217 & 0.060 & 0.9997 \\
& 5.0 & 0.9922 & 0.195 & 0.056 & 0.9997 \\
& 10.0 & 0.9919 & 0.134 & 0.055 & 0.9994 \\
& EDL & 0.9912 & 0.196 & 0.054 & 0.9996 \\
& RED & 0.9921 & 0.196 & 0.048 & 0.9996 \\
\midrule
\multirow{7}{*}{SVHN} 
& 0.1 & 0.9130 & 0.407 & 0.157 & 0.9833 \\
& 1.0 & 0.9130 & 0.409 & 0.153 & 0.9840 \\
& 2.5 & 0.9097 & 0.308 & 0.153 & 0.9833 \\
& 5.0 & 0.9128 & 0.213 & 0.150 & 0.9847 \\
& 10.0 & 0.9127 & 0.124 & 0.154 & 0.9750 \\
& EDL & 0.8669 & 0.116 & 0.167 & 0.9737 \\
& RED & 0.8678 & 0.118 & 0.164 & 0.9743 \\
\midrule
\multirow{7}{*}{CIFAR-10} 
& 0.1 & 0.7250 & 0.415 & 0.252 & 0.9055 \\
& 1.0 & 0.7176 & 0.413 & 0.255 & 0.8996 \\
& 2.5 & 0.7133 & 0.315 & 0.250 & 0.8954 \\
& 5.0 & 0.7133 & 0.205 & 0.248 & 0.8899 \\
& 10.0 & 0.7126 & 0.128 & 0.244 & 0.8803 \\
& EDL & 0.5968 & 0.131 & 0.294 & 0.8128 \\
& RED & 0.6024 & 0.111 & 0.293 & 0.8213 \\
\bottomrule
\end{tabular}
}
\end{table}
\begin{table*}[t]
\centering
\caption{Class distribution and semantic labels in CIFAR-10-LT.}
\label{tab:class_dist}
\begin{tabularx}{\textwidth}{l|XXXXXXXXXX}
\toprule
\textbf{Class Index} & 0 & 1 & 2 & 3 & 4 & 5 & 6 & 7 & 8 & 9 \\
\midrule
\textbf{\# Samples} & 5000 & 2997 & 1796 & 1077 & 645 & 387 & 232 & 139 & 83 & 50 \\
\textbf{Label} & Airplane & Auto & Bird & Cat & Deer & Dog & Frog & Horse & Ship & Truck \\
\bottomrule
\end{tabularx}
\end{table*}
\subsection{Additional Details: Category Imbalance Experiments}

To evaluate the robustness of our method under class imbalance, we conduct experiments on the CIFAR-10-LT dataset. This benchmark is derived from CIFAR-10 by applying an exponential decay function to the class sample sizes, producing a long-tailed distribution that reflects real-world data imbalance..

The resulting dataset exhibits a pronounced imbalance ratio between head and tail classes. As shown in Table~\ref{tab:class_dist}, Class 0 (Airplane) contains 5000 samples, whereas Class 9 (Truck) contains only 50, forming an imbalance ratio of 100:1. Such a distribution presents challenges for both classification accuracy and uncertainty calibration in minority classes.


\end{document}